\documentclass{article}

\usepackage{arxiv}
\usepackage[utf8]{inputenc} 
\usepackage[T1]{fontenc}    
\usepackage{hyperref}       
\usepackage{url}            
\usepackage{booktabs}       
\usepackage{amsfonts}       
\usepackage{nicefrac}       
\usepackage{microtype}      
\usepackage{lipsum}
\usepackage{xcolor}
\usepackage{graphicx}

\usepackage{algorithmic}
\usepackage{algorithm}
\usepackage{amssymb,amsmath}
\usepackage{array}

\title{An Evolutionary Algorithm of Linear complexity: Application to Training of Deep Neural Networks}

\author{ S.~Ivvan Valdez\\
  Metropolitan Observatory\\
  CONACYT Research Fellow\\
  Quer\'etaro, Qro., Mexico\\
  \url{sergio.valdez@conacyt.mx}
\And
 Alfonso Rojas-Dom\'inguez \\
 Posgraduate \& Research Division\\ Le\'on Institute of Technology\\
 Av. Tecnol\'ogico SN, Le\'on, 37290, Mexico \\
 \url{alfonso.rojas@gmail.com} 
}

\begin{document}

\maketitle

\begin{abstract}
The performance of deep neural networks, such as Deep Belief Networks formed by  Restricted Boltzmann Machines (RBMs), strongly depends on their training, which is the process of adjusting their parameters.
This process can be posed as an optimization problem over $n$ dimensions. However, typical networks contain tens of thousands of parameters, making this a  High-Dimensional Problem (HDP).
Although different optimization methods have been employed for this goal, the use of most of the Evolutionary Algorithms (EAs) becomes prohibitive due to their inability to deal with HDPs. For instance, the Covariance Matrix Adaptation Evolutionary Strategy (CMA-ES) which is regarded as one of the most effective EAs, exhibits the enormous disadvantage of requiring $O(n^2)$ memory and operations, making it unpractical for problems with more than a few hundred variables.
In this paper, we introduce a novel EA that requires $O(n)$ operations and memory, but delivers competitive solutions for the training stage of RBMs with over one million variables, when compared against CMA-ES and the Contrastive Divergence algorithm, which is the standard method for training RBMs.
\end{abstract}

\keywords{Evolutionary Algorithm \and Variance Direction \and Deep Learning \and Deep Neural Networks \and High Dimensional Optimization}

\section{Introduction}

Deep Neural Networks (DNN) are widely used for classification tasks, with impressive performance in contrast with other methods and even against human experts~\cite{lecun2015deep}. Every neural model contains a number of parameters (traditionally known as weights and biases) that must be adjusted so that the model performs the desired task correctly. The process of adjusting the network parameters is known as \emph{training}. The training of a DNN is modelled as an optimization problem, that is to say, the minimization of an error function computed over the training data instances. Hence, the performance of a DNN completely depends on the training stage.

As a rule, the solution of the optimization problem has been approximated via gradient-based (GB) and quasi-Newton optimizers~\cite{bottou:2018}, despite the fact that the problem presents multiple minima where such optimizers can be trapped. In addition, each kind of DNN requires of an specific training method because the GB optimizers explicitly compute derivatives of the parameters, including the specific DNN architecture and the transfer and activation functions. Even more, the \emph{Fundamental Deep Learning Problem of Gradient Descent}~\cite{SCHMIDHUBER201585} establishes that some DNN architectures cannot be efficiently trained by gradient-based optimizers via backpropagation, due to the \emph{gradient vanishing problem}. In consequence, further research in alternative training methods is crucial for continuing expanding the current knowledge.

In contrast with the standard GB approaches, Evolutionary Algorithms (EAs) and other population-based stochastic methods exhibit two advantages: first, that these can escape from local minima, and second that these do not require of problem-specific information (such as first or second derivatives), hence, one EA method could be used for a variety of DNNs by virtue of a complete decoupling from the DNN architecture and transfer- and activation-functions used throughout the network.

On the other hand, a great disadvantage of EAs is their computational cost, in memory as well as operations. The adequate population size for most EAs is dependent on the number of optimization variables~\cite{mora2017population}, often in a superlinear proportion. Some researchers have proposed EAs which have been successfully used for solving high-dimensional problems~\cite{loshchilov2013cma} using a population size which linearly depends on the number of variables~\cite{hansen2003reducing}.  Nevertheless, EAs in general have not been adopted for training or pretraining of DNNs because of their computational cost, which makes these unpractical against the current approaches based on other paradigms~\cite{ZHANG2016146} despite their competitive results~\cite{petrosoky:deepneu}.

There exist evolutionary algorithms applied to High Dimensional Problems (HDPs) \cite{deb2016breaking}, but they have not been applied to DL, or to highly non-linear optimization problems, such as the one attempted here. Additionally, other authors intended to use descent directions in population based algorithms \cite{SierraUrrecho2008}, although, these proposals have not reached the deep learning field, mainly because they are designed and tested for low-dimension problems.

\section{From the $O(n^2)$ CMA-ES to the design of an $O(n)$ algorithm }

Evolutionary Algorithms have been widely used for addressing non-derivable problems with multiple optimal solutions~\cite{Cheng2016}. In this context, the most effective algorithm reported in the literature for high-dimensional problems in continuous domains is the Covariance Matrix Adaptation Evolutionary Strategy (CMA-ES)~\cite{loshchilov2013cma}.

The CMA-ES uses the covariance matrix for biasing the search towards promising directions. Although it delivers the most competitive approximations to the optima (even compared with quasi-Newton algorithms for convex problems such as Levenberg-Maquard and BFGS~\cite{loshchilov2013cma}), the CMA-ES requires of $O(n^2)$ memory and operations because it computes a covariance matrix oriented towards promising directions computed from the historical trajectory of the elite solution; this requires of $n^2$ memory locations for $n$ optimization variables. Furthermore, sampling from a multivariate normal distribution with a full covariance matrix requires factorizing it or computing its eigenvectors, which requires of $n^2$ more memory locations and operations. 

Translating these observations to the context of training DNNs, where the number of variables to be optimized is in the range $10^5-10^7$, the covariance matrix storage and factorization would require at least $2\times 10^{10}$ memory locations. Assuming a double-precision floating-point format (64 bits or 8 bytes) for storage, then the training of a small DNN with CMA-ES would occupy about 160 GB of RAM.

Even though the CMA-ES is unpractical for high dimensional optimization problems such as those encountered in DNNs training, it possesses relevant features which could be included in the design of a $O(n)$ algorithm. 

\subsection{Previous work}
It is known that the CMA-ES main steps are quadratic, while some other, for instance the sampling method, is cubic \cite{ros2008simple}; this elevated computational complexity is hinders the application of CMA-ES to high-dimensional problems(HDPs). In consequence, the problem of designing a linear algorithm that includes the core ideas of the CMA-ES has been attempted before, prominently by Loshchilov~\cite{Loshchilov:2014}, Ros and Hansen \cite{ros2008simple},  and Arnold and Hansen \cite{arnold2010active}.

In one of Loshchilov's previous proposals \cite{Loshchilov:2014}, the algorithm computes the same trajectories as the CMA-ES and then updates a Cholesky factorization of the covariance matrix containing the most relevant dimensions. Nevertheless, the author intends to preserve as most as possible the CMA-ES structure and work-flow.

A similar scheme which reduces the cubic steps to quadratic complexity was presented by Arnold and Hansen~\cite{arnold2010active}; they employ a $(1+1)$ strategy for the adaptation of the covariance matrix, reducing the computational cost. These previous proposals are well founded and have reported adequate results, but they are far from a competitive approach to be used in Deep Learning (DL), considering that the largest number of dimensions that have been discussed in those works is $n=40$, while the optimization problems in DL are at least two orders of magnitude greater. 

In Ros and Hansen's proposal~\cite{ros2008simple} a diagonal matrix is used to achieve linear complexity. This approach is intended for separable objective functions, and in our opinion loses one of the main advantages of CMA-ES, which is the orientation of the covariance matrix of the population of solutions towards a promising direction.

Contrary to these previous approaches, in this work we develop a naturally linear algorithm without the aim of preserving the CMA-ES work-flow but still allowing it to serve as an inspiration for our proposal. Namely, we introduce a novel Evolutionary Algorithm based on two proposals: a search based on a diagonal covariance matrix  (that requires $O(n)$ operations and memory), of which the main variance direction is aligned with a \emph{single} promising direction that in turn is computed as a combination of two directions derived from the evolution of the best individual through previous iterations. The storage and computation requirements for all of these operations are $O(n)$.  

\section{The Linear Evolutionary Algorithm based on the Main Variance Directions (LEA-MVD) }
\label{sec:MVDAA}

The main idea of our proposed algorithm is as follows. After proper initialization and evaluation of a population of solutions (discussed below), the algorithm is expected to produce new candidate solutions based on a guided search. Three main components are used for guiding this search: the first component is a Normal distribution, which models the most promising region within the solution space; the second component is a direction based on the evolution of the best individual trough generations; the third component is a direction based on the descent from the worst individuals to the best individual in the current population. The three components are used for sampling new solutions within a promising region and to translate that sample in a potentially descent direction. Our proposal is thus termed Linear Evolutionary Algorithm based on the Main Variance Directions (LEA-MVD) and is shown in Algorithm \ref{alg:LEA-MVD}.

	\begin{algorithm}[H]
	 	\begin{algorithmic}[1]
	 	     	\footnotesize
	 	     \REQUIRE 
	 	     $\lambda$: population size,
	 	     $n_{elite}$: number of elite individuals,             
	 	     $n_{var}$: number of optimization variables,
	 	     $x_{inf},x_{sup}$: superior and inferior search limits,
	 	     $n_{gen}$: stopping criterion (when no-convergence is reached),
	 	     $\sigma_{min}$: stopping and convergence criterion.
			 \STATE $t=1$
			 \STATE $X^t=$ InitialPopulation($\lambda,n_{var},x_{inf},x_{sup}$).
			 \STATE $F^t=$Evaluation$(X,f())$ .\\
			 \STATE $[ix,ibest]=$Selection($F^t,nelite$).\\
			 \STATE $X_{elite}=X^t(ibest,)$, $F_{elite}$=$F^t(ibest)$, $f^t_{best}=F_{elite}(1)$,  $x^{t-1}_{best}=x^t_{best}=X_{elite}(1,)$,  $count_{felite}$, $\beta_1=1$, $\beta_2=0.9$, $\Sigma^t=diag(\sigma_{min}+1)$.
			 \WHILE {$t<n_{gen}~~\&~~ norm(\Sigma^t)<\sigma_{min}$ }
			    \STATE $[\mu^t, \Sigma^t]=$
			    MeansAndStandardDeviation($X^t,F^t,ix$)
			    \STATE $[C^t_{ani},\mu^t_{ani},\sigma^t_{ani},P^t]=$MainDirectionEstimators($X,ix,C^{t-1}_{ani},P^{t-1},x^t_{best},x^{t-1}_{best}$)
                \IF{$count_{felite}=10$} 
                \STATE {$\Sigma^t=\{1\}^{n_{var}}$, $count_{felite}=0$, $\beta_1$=0.1}
                \ENDIF
                \STATE $X^{t+1}$=[$X^t_{elite}$,RePopulate($\lambda-n_{elite},\mu^t,\Sigma^t,C^t_{ani},\sigma^t_{ani},P^t, \beta_1,\beta_2$)]
                \STATE $t=t+1$.
			    \STATE $F^t=$Evaluation$(X,f())$ .\\
			    \STATE [$ix,ibest$]=Selection($F^t,nelite$).\\
			    \STATE $X_{elite}$=$X^t(ibest,)$, $F_{elite}$=$F^t(ibest)$. $f^t_{best}=F_{elite}(1)$,  $x^t_{best}=X_{elite}(1,)$.
                \IF{$f_{best}^{t-1}\neq f_{best}^{t}$}{
                  \STATE $count_{felite}=0$.
                  \STATE $\beta_2=min(1,0.2+\beta_2)$
                  \IF{$\beta_1>1$}
                    {\STATE $\beta_1=min(3,1.1\beta_1)$}
                    \ELSE{
                    \STATE $\beta_1=1.4\beta_1$
                   }\ENDIF
                }\ELSE{
                \STATE $count_{felite}=count_{felite}+1$
                \STATE $\beta_2=max(0,\beta_2-0.1)$
                \IF{$\beta_1<1$}{\STATE $\beta_1=0.8\beta_1$ }\ELSE{\STATE $\beta_1=0.5\beta_1$}\ENDIF
                }            
                \ENDIF
			 \ENDWHILE
			 \STATE Return $[x^t_{best},f^t_{best}]$
		 \end{algorithmic}
		 \caption{The Linear Evolutionary Algorithm based on the Main Variance Directions (LEA-MVD).}\label{alg:LEA-MVD}
	\end{algorithm}

The LEA-MVD requires of two sensitive parameters: the population size, $\lambda$, and the number of elite individuals, $n_{elite}$. Even they are considered sensitive for getting the best performance, a sufficient competitive performance can be achieved with fixed values, in particular the number of elite individuals can be fixed to $n_{elite}=4$. For the preliminary results presented here the population size is fixed to $\lambda=20$, although a more deeper analysis must be conducted for determining a better rule for setting these parameters. 

\subsection{Implementation Details}

\paragraph{Input Parameters.--\hspace{-0.5em}} the input parameters are the population size $\lambda$ (fixed to 24 for these preliminary results); the number of elite individuals $n_{elite}=4$ (this fixed value is suggested for any problem); the number of optimization variables $n_{var}$; and the inferior limit $x_{inf}$ and superior limit $x_{sup}$ of the solution space, which in general are defined by the problem. For RBM pretraining, we suggest a symmetric range around 0, inside $[-1,1]$ for all variables, in the experiments we employed $[-0.1,0.1]$.

\paragraph{Stopping criteria.--\hspace{-0.5em}} the algorithm is stopped if the norm of the standard deviations is less than $\sigma_{min}=10^{-4}\sqrt{n_{var}}$ or if the number of generations $n_{gen}=30$.

\paragraph{Initial Population.--\hspace{-0.5em}} we test two initialization procedures to obtain the initial population $X^1$, which is a $\lambda \times n_{var}$ matrix: (A) by uniformly sampling between $x_{inf}$ and $x_{sup}$; (B) by sampling from a Normal distribution $\mathcal{N}(\mu=seed,\sigma=0.1)$, where seed is an initial solution, this case is discussed in the Results section.

\paragraph{Evaluation.--\hspace{-0.5em}} the inputs for the evaluation are the current population $X^t$ and the objective function $f(\cdot)$. This procedure returns the objective value as an one-dimensional array $F^t$ for iteration $t$. In our experiments, each individual is a set of parameters of an RBM, hence the evaluation is carried out by computing the reconstruction error of the input-output data as our objective function.

\paragraph{Selection.--\hspace{-0.5em}} the selection returns two index subsets, $ibest$ and $ix$, the second is the subset of indexes of the decreasing order of the population $F^t$, while $ibest$ are the first $n_{elite}$ indexes of $ix$.

\subsection{First Component: Sampling from the most promising region}\label{sec::firstComponent}
A probabilistic model is used to identify the most promising region, so that the algorithm can sample from it with a high probability. This is achieved by means of the Empirical Selection Distribution (ESD)~\cite{valdez2009approximating} which consists of assigning a set of discrete weights $G_{ix}$ to each solution $ix$, according to Eq.~\eqref{eq:fractour}. These weights can be seen as probabilities or relative frequencies associated to each individual (the better an individual, the larger its weight).  

According to Valdez~\cite{valdez2009approximating}, sampling from the ESD is equivalent to applying a selection operator. For instance, if the exponent in Eq.~\eqref{eq:fractour} were $1$, this distribution would be equivalent to a binary tournament. If the exponent were $2$, the distribution would be equivalent to a double binary tournament (compute a selected set by a binary tournament, then apply another binary tournament over the first selected set). In our case, we choose to use a \emph{fractional tournament} with exponent equal to $1.5$, which according to prior experimentation, adequately modulates the selection pressure.\vspace{-1em}

\begin{eqnarray}\label{eq:fractour}
    G_{ix}=\frac{\hat G_{ix}^{1.5}}{\sum(\hat G^{1.5}_{ix})},\\
    \nonumber
    \text{where } \hat G_{ix}=\{\lambda,\lambda-1, \lambda-2,\ldots,1\}
\end{eqnarray}

The Normal model is defined by a set of means and standard deviations, hence it could be seen as a multivariate Normal distribution with a diagonal covariance matrix. In other words, the random variables are considered independent. Thus, the mean and standard deviations are computed for the $i$-th variable according to Eqs. \eqref{eq:mean} and \eqref{eq:sigma}, respectively.\vspace{-1em}

\begin{eqnarray}
    \label{eq:mean}
    \mu_i= \sum^{\lambda}_{ix=1} X_{ix,i} \cdot G_{ix}\\
    \label{eq:sigma}
    \Sigma_{i,i}= \sqrt{\sum^{\lambda}_{ix=1} (X_{ix,i}-\mu_i)^2 G_{ix}}
\end{eqnarray}

\subsection{Second and Third Components: Estimators of the Main Variance Direction}

One of the main contributions of this proposal is to use a $rank(1)$ covariance matrix instead of a $rank(n)$. The latter can be decomposed in $n$ eigenvectors associated with $n$ eigenvalues, the greatest of which corresponds to the largest variance direction. In this context, our proposal consists in setting the Main Variance Direction (MVD) as the direction of greatest improvement, and using this single direction to translate the population of candidate solutions. For this purpose we compute two descent directions, named as the second and third components which, as explained above, are combined into a single direction. The first descent direction is given by the improvement of the elite individual, while the second is given by the descent vector from the worst solutions in the population to the best. Possibly, both directions are similar, hence it is not desirable nor necessary to perform two translations in the same direction. In consequence, we propose to compute the first direction and then to compute from the second direction only the part that is orthogonal to the first. That is to say, the descent vector from the worst solutions to the best that is orthogonal to the improvement of the elite individual.

The first direction $P^t$ is related to the improving path of the best solution. It is computed as $P^t=0.1(x^t_{best}-x^{t-1}_{best})+0.9P^{t-1}$, notice that it is a combination of the vector from the best solution at prior iteration $t-1$ to the best solution at current iteration $t$, and the direction $P^{t-1}$, from the prior iteration.

The second descent direction is obtained  by computing the vector of maximum projection to the matrix $U=[X(ix_1,:)-X(ir_1,:),X(ix_1,:)-X(ir_2,:),X(ix_1,:)-X(ir_3,:)]$, subject to it being orthogonal to $P^t$. Notice that $ix$ are the population indexes decreasingly sorted, thus $X(ix_1,:)$ is the elite individual and $ir$ represents a set of indexes randomly taken from $\{ix_2,ix_3,...,ix_\lambda\}$. In other words, we select indexes from the $\lambda-1$ worse individuals in the population. For the experiments in this paper $|ir|=4$.

The vector of the maximum projection, called anisotropic descent direction $\hat C^t_{ani}$, can be computed using the power method over $U$, removing the projection to $P^t$ at each iteration. Finally, the second direction, is $C^t_{ani}=0.1\hat C^t_{ani} + 0.9C^{t-1}_{ani}$. After computing the $C^t_{ani}$  direction, we compute the mean $\mu_{ani}$ and standard deviation $\sigma_{ani}$, of the projections of the vectors $[X(ix_2,:)-X(ix_1,:),X(ix_2,:)-X(ix_1,:),...,X(ix_\lambda,:)-X(ix_1,:)]$ over $C^t_{ani}$. The quantities $\mu_{ani}$ and $\sigma_{ani}$, are used for translating the sampled points in this direction.

\subsection{Regenerating the Population}

Our algorithm is equipped with elitism, so that at each iteration the best $n_{elite}$ individuals are preserved and $\lambda-n_{elite}$ individuals are regenerated. The sampling of the new population is as follows:

\begin{enumerate}
\item $\lambda-n_{elite}$  samples for $X$ are generated by using univariate Normal distributions with mean=$0$ and the standard deviations computed according to the first component, described in Section \ref{sec::firstComponent}.
\item Then, a direction vector which combines the second and third components (first and second directions) is computed using Eq.~\eqref{eq:searchdir}. This direction is used for shifting $X$ in a promising direction.
    \begin{equation}
        \label{eq:searchdir}
        v=\beta_2P^t+(1-\beta_2)\mu_{ani}C^t_{ani}
    \end{equation}
    Notice that $\beta_2$ regulates which direction is the dominant. 
    \item For each dimension $i$, that is a column of the sample $X$, Equation \eqref{eq:sampling}
    is used for generating the new sample. Notice that $\beta_1$, regulates the shift inserted by direction $v$. 
    
    \begin{equation}
    \label{eq:sampling}
     \hat X_i= (X_i+\mu_i+\beta_1v_i)
    \end{equation}    
    \item Finally, with a small probability of 0.02, a dimension could be enlarged or shortened as follows:
    
       \begin{equation}
        factor=
        \begin{cases}
            1+\mathcal{U}(-0.5, 0.5) & $if $\ \mathcal{U}(0, 1)<0.02\\
            1 & $otherwise$
        \end{cases}
    \end{equation} 

    so that,  $\bar X_i= (factor)\hat X_i$. This last step is a perturbation in 2\% of the dimensions.
    
    The union of $\bar X$ and the $n_{elite}$ individuals preserved from the prior population form the new population.
\end{enumerate}

\paragraph{The step sizes $\beta_1$ and $\beta_2$.--\hspace{-0.5em}} the parameter $\beta_1$ regulates the total descent; while the best solution is improved (signaling that the descent direction is adequate) it increases, otherwise it is reduced. Meanwhile, $\beta_2$ regulates the dominant direction among $P^t$ and $C_{ani}$; if there is not a descent (meaning that the elite solution is stagnated), the weight of $C_{ani}$ is increased, and vice versa.

\section{Results}
In this section we contrast the results obtained for the pretraining of a Deep Belief Network (DBN) using three algorithms: the LEA-MVD (our proposal); the CMA-ES, which is considered the most competitive in a set of test poblems~\cite{CMAES:Online}, \cite{Hansen}; and Contrastive Divergence (CD), which is the predominant algorithm for pretraining DBNs.

\subsection{Experimental setup}
\begin{figure}[!h]
\centering
\includegraphics[scale=0.4]{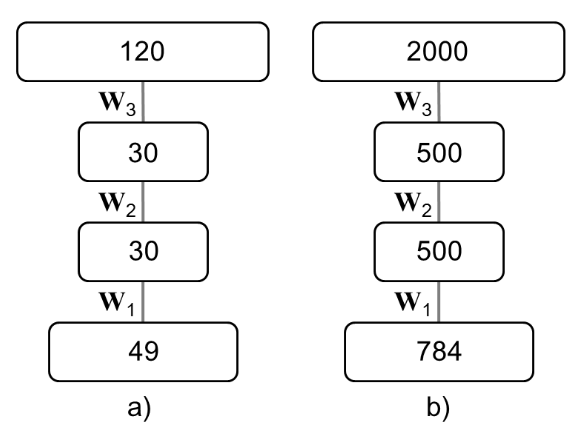}
\caption{DBN architecture used for (a) images of $7\times7$ pixels. (b) images of $28\times28$ pixels. The quantities in the boxes indicate the number of nodes in each layer of a RBM (two layers per RBM). Three stacked RBMs form a DBN.}
\label{fig:Architecture}
\end{figure}

A DBN is formed by stacking together several RBMs as illustrated in Fig.~\ref{fig:Architecture} (we use a similar architecture to that introduced by Hinton and Salakhutdinov~\cite{hinton2006reducing}), where $W_j$ represents the weight matrix (the matrix of parameters) of the $j$-th RBM. The DBN used in our experiments is intended to perform handwritten-digit classification. To evaluate the performance, the MNIST (\textit{Modified National Institute of Standards and Technology}) database of handwritten digits was employed. This database consists of binary images: $60,000$ for training and $10,000$ for  testing; each image is of size $28 \times 28$ pixels. 

Notice that the pretraining of, for instance, the parameters in $W_1$, involves the optimization of approximately $n = 784 \times 500=392,000$ variables. Since the CMA-ES requires of $O(n^2)$ memory allocations, this implies the use of several hundred GB of storage for a single matrix, which is prohibitive for a workstation. Therefore, in order to compare our approach against the CMA-ES, the images in the MNIST database were scaled down to 25\% of their original size (i.e. to $7 \times 7$ pixels). 

The training of a DBN implies pretraining each of the RBMs in its architecture sequentially but independently. The number of trainable variables in the $k$-th RBM is cmputed as $|W_k|=v_j+h_j+v_j+h_j$, where $v_j$ and $h_j$ represent the number of nodes in the visible (input) and hidden layers of the RBM, correspondingly. Since our DBN is formed by three RBMs, then we have three optimization problems. For the training based on scaled down images (Fig. \ref{fig:Architecture}-a), these problems involve: $|\mathbf{W}_1|=1,549$ variables, $|\mathbf{W}_2|=960$ variables and $|\mathbf{W}_3|=3,750$ variables.

In a second experiment, the LEA-MVD is compared against CD. Since the CMA-ES is not involved in this experiment, the related limiting memory requirements do not apply, so that the training images were used in their original resolution. For the training using the images with their original size (Fig. \ref{fig:Architecture}-b), the number of optimization variables are: $|\mathbf{W}_1|=393,284$ variables, $|\mathbf{W}_2|=251,000$ variables, and $|\mathbf{W}_3|=1,002,500$ variables.

The CD algorithm typically used to perform the pre-training of each RBM converges within 50 iterations  (thus the complete pre-training of the DBN takes 150 iterations). Because of this, the algorithms that were compared against CD were allocated the same number of iterations.

\subsection{Experimental Results of LEA-MVD vs CMA-ES and CD}

\begin{figure}[hbt]
\centering
\includegraphics[width=0.9\linewidth]{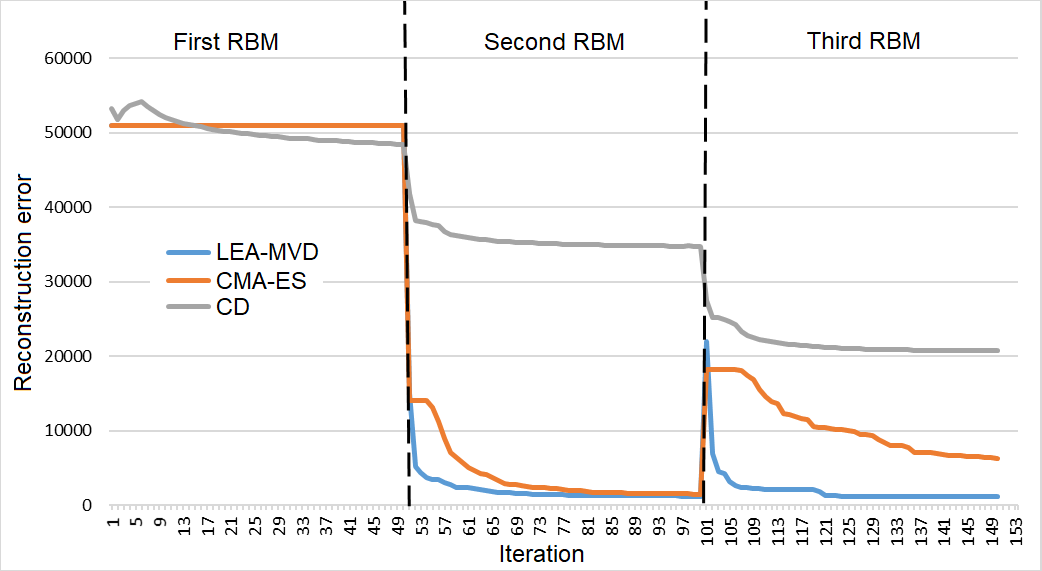}
\caption{Reconstruction error obtained by LEA-MVD, CMA-ES and CD for DBN training with images of $7 \times 7$ pixels.}
\label{fig:reconstruction-error}
\end{figure}

\paragraph{First Experiment.--\hspace{-0.5em}} the reconstruction error for the three algorithms compared under the training scheme using the scaled-down version of the training data is shown in Fig.~\ref{fig:reconstruction-error}.  The plots illustrate a typical execution of the three algorithms. The CMA-ES and the LEA-MVD were initialized using as seed solution that delivered by CD at the first iteration. Even though this seed solution could not be improved for the first RBM, for the second and third RBMs the CMA-ES and LEA-MVD achieved significantly smaller reconstruction errors than the widely-used CD, with LEA-MVD being the best of the three. Hence, our proposal not only requires much lesser memory and operations but it also delivers better solutions than the CMA-ES and CD.

\begin{figure}[hbt]
\centering
\includegraphics[width=0.9\linewidth]{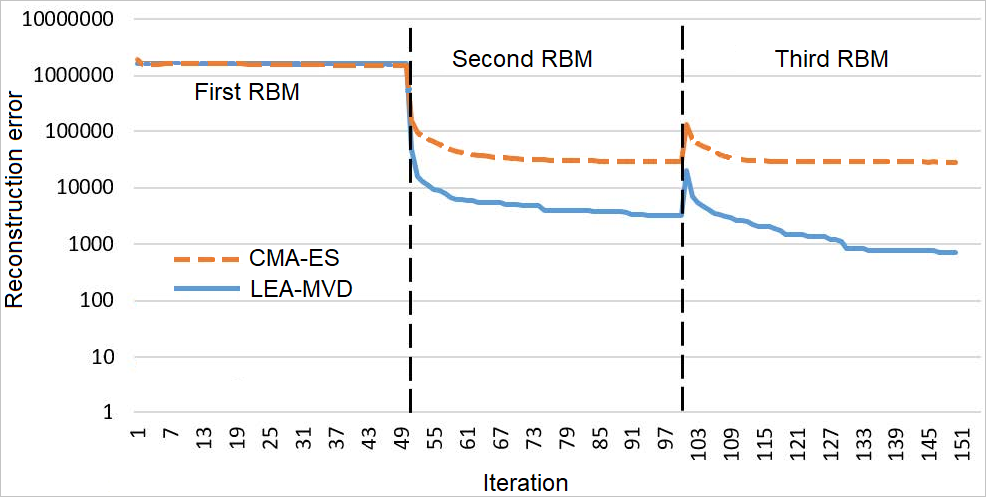}
\caption{Reconstruction error delivered by LEA-MVD and CD for the original sized images of $28\times 28$ pixels.}
\label{fig:errorPlot6}
\end{figure}

\paragraph{Second Experiment.--\hspace{-0.5em}} after showing in the above comparison that LEA-MVD can overcome the performance of CMA-ES in the limit imposed by its memory requirements, a second experiment was conducted to compare LEA-MVD against CD using the images in their original size of $28\times28$ pixels. This second experiment was executed 30 times in order to provide statistical robustness. The average result of these 30 executions, per iteration, is shown in Fig.~\ref{fig:errorPlot6}. 

As can be appreciated, the reconstruction error in this second experiment is generally larger than that in the first experiment. This is due to the larger size of the images used for the training of the DBN. Notice as well that after an initial improvement of the error (barely noticeable in the far right of the plots), neither of the algorithms managed to find a significantly better solution for the first RBM. For the second RBM, CD converged to a solution corresponding to a reconstruction error of about 50,000 (notice that the error is shown in logarithmic scale), while LEA-MVD produced a solution that is about one order of magnitude better, with an error of approximately 5,000. For the third RBM, CD produced only small improvements that cannot be appreciated at the scale shown. In contrast, LEA-MVD further improved the solution from the previous RBM and manage to reduce the reconstruction error to approximately 900.

These results unequivocally illustrate that the LEA-MVD algorithm is significantly superior than CD for the task of pre-training of a DBN.

\section{Conclusions}

In this paper we introduced a novel evolutionary algorithm based on the paradigm of Estimation of Distribution Algorithms which has $O(n)$ complexity in memory and operations. Our proposal uses two probabilistic components and one deterministic component for sampling candidate solutions. The first probabilistic component estimates the position of a promising region, while the second models promising directions: the direction taken from the improvement of the elite individual is a single value, and the direction taken from the direction from the worse solutions to the best solution in the current population is used for computing the second probabilistic component, which is a univariate Normal distribution over the direction $C_{ani}$.

In all cases the updating rules consider fixed learning rates, that is to say, the current directions depend on the current information as well as on the historical information; this provides robustness to the estimation of the directions in the same way that stochastic gradient does \cite{kingma2014adam}.

The algorithm introduced herein is competitive with two state of the art algorithms, the CMA-ES, which is regarded in the literature as one of the most competitive evolutionary algorithms for continuous optimization, and Contrastive Divergence, which is the default algorithm for pretraining of RBMs / DBNs. 
The main contribution of this work is the combination of different paradigms, namely the Estimation of Distribution Algorithms with improving paths (akin to the CMA-ES), and robust estimators similar to  Stochastic Gradient methods. All of this, results in a competitive optimization algorithm, as has been illustrated for the problem of pre-training a deep neural network. 

Furthermore, the design of LEA-MVD includes self-adaptive parameters, this is an attractive feature that enables practitioners to avoid the problem of hyper-parameter tuning. Future work contemplates the analysis of the performance of this proposal for larger deep learning problems, in order to improve their operators and to reduce, even more, the computational cost.

\paragraph{Acknowledgement.} This work was supported by the National Council of Science and Technology of Mexico (CONACYT), through Research Grants: C\'ATEDRAS-7795 (S.I. Valdez) and C\'ATEDRAS-2598 (A. Rojas).  

\bibliographystyle{plain}
\bibliography{references.bib}
\end{document}